\documentclass[12pt,a4paper]{article}
\usepackage[left=2cm, right=2cm]{geometry}
\usepackage{setspace}
\usepackage[utf8]{inputenc}
\usepackage[utf8]{inputenc}
\usepackage{graphicx}
\graphicspath{ {images/} }
\usepackage{amsmath}
\usepackage{wrapfig}
\usepackage{geometry}
\usepackage[normalem]{ulem}
\usepackage{booktabs}
\usepackage{subcaption}
\usepackage{natbib}
\setcitestyle{numbers,open={(},close={)}}

\usepackage{parskip}
\usepackage{pifont}
\usepackage{amsfonts}
\usepackage{amssymb}
\usepackage{longtable}
\usepackage{xcolor}
\usepackage{multicol}
\usepackage{placeins}
\usepackage{tcolorbox}
\usepackage{soul} 
\usepackage{fvextra}

\title{AI evaluation may bias perceptions \\
\Large{The importance of context in interpreting academic writing}
}

\author{
\medskip
 Shang Wu\\ UC Irvine \\ shangw13@uci.edu \\
        \and 
       Randol Yao\\\\[-1.9ex] MIT \\ hyyao@mit.edu
}

\date{May 19, 2026}

\begin{document}
\maketitle
\begin{abstract}
This paper examines how estimates of AI use in scientific writing can be biased when evaluation methods ignore contextual differences across countries and fields. Using large-scale data on journal publications from Dimensions, we construct AI-likeness benchmarks based on differences between human-written and LLM-rephrased abstracts. We show that a pooled benchmark may confound pre-existing stylistic variation with AI-generated text, producing substantial distortions across country-field groups even in pre-LLM publications. In contrast, country-field-specific benchmarks attenuate such distortions and provide a more credible baseline for comparison. Applying these methods to publications in 2025 reveals that the pooled benchmark systematically overestimates AI use in certain countries and fields while underestimating it in others. These findings highlight the importance of context-aware measurement for accurate and equitable evaluation of AI use in science.
\end{abstract}
\vspace{1\baselineskip}

\begin{center}
\textit{“You should not use em dashes—that’s a sign of AI-generated writing.”} \\[4pt]
\textit{“But I’ve been using them long before AI...”}
\vspace{1\baselineskip}
\end{center}

\setstretch{1.5}

What scientific knowledge gets recognized depends not only on its intrinsic merit but also on how that knowledge is perceived and evaluated. The valuation of scientific knowledge can vary based on the gender composition of the research team \citep{subramani2025investigating} and across researchers' countries of origin \citep{qiu2025stands}. Such observable cues are often used as ``shortcuts" for quality, especially under limited expertise, information, or time pressure \citep{fry2024author}. And the recent rise of large language models (LLMs) in scientific writing introduces a new layer to this problem. \citet{kusumegi2025scientific} reports that LLM adoption in scientific writing is associated with increases in scientific productivity, especially for non-native English speakers, while simultaneously eroding traditional quality signals like writing complexity. Other research also attempts to estimate the use of LLMs in science and how it affects research direction and scientific governance \cite{hao2026artificial, he2026academic}.

Importantly, these estimates of LLM use in scientific writing often rely on a strong assumption: that “AI-likeness” can be measured against a common benchmark and meaningfully compared across articles produced in different countries and fields. Such an assumption is consequential because, in reality, scientific writing does not follow a single global standard: longstanding differences in linguistic convention, disciplinary norms, and rhetorical style mean that some countries and fields may appear more “AI-like” than others even before the rise of LLMs. When a single standard is applied across such heterogeneous contexts, differences in writing style can be mistaken for differences in AI use. This is important because readers tend to discount writing perceived to be AI-generated \cite{raj2026artificial}, so misclassifications can introduce systematic bias across countries and fields, and further reinforce existing inequalities in scientific evaluation \cite{subramani2025investigating, qiu2025stands, fry2024author}.

This paper argues that accurate evaluation of AI use in scientific publications requires accounting for country-field-specific contexts. We demonstrate this by comparing estimates of AI-like text share with and without accounting for such heterogeneity. Specifically, we retrieve metadata for all English-language journal publications from Dimensions and define 18 country groups (hereafter, “countries”) and 13 fields of study: Appendix Tables B1 and B2 list the corresponding classifications. From each of the 234 country-field groups, we draw up to 2,000 publications from 2020. We then use \textit{ChatGPT-4o-mini}, a widely-used large language model, to rephrase the abstracts of all sampled publications following a two-step procedure adapted from \citet{kusumegi2025scientific} and \citet{liang2025quantifying}. The resulting pairs of original and AI-rephrased abstracts form a reference corpus for estimating word distributions characteristic of human-written and AI-rephrased scientific text. We then tokenize the original and rephrased texts, compute word-level log-odds ratios, and construct two sets of AI-likeness benchmarks: (i) a pooled benchmark trained on all sampled publications, as in prior work\footnote{Prior work fits different models for each pre-print source, e.g., \textit{arXiv}, \textit{bioRxiv}, \textit{SSRN} \cite{kusumegi2025scientific, he2026academic, liang2025quantifying}. In this paper, we use journal publications from a single source.}; and (ii) 234 country-field-specific benchmarks, each trained only on publications within the corresponding country-field group. Here, a ``benchmark" refers to a list of words and the corresponding log-odds ratios for each word, where the log-odds ratio captures how much more or less likely the word is to appear in human-written versus AI-rephrased abstracts. Specifically, under the pooled benchmark, the log-odds ratio of word $t$ is
\[
\text{log-odds ratio}_t = \log\left(\frac{P_t/(1-P_t)}{Q_t/(1-Q_t)}\right),
\]
where $P_t$ and $Q_t$ are the pooled probabilities that word $t$ occurs in the actual human-written abstract and the AI-rephrased abstract, respectively. More negative values indicate words appeared relatively more often in AI-rephrased abstracts. By contrast, the country-field-specific log-odds ratio is
\[
\text{log-odds ratio}^g_t
=
\log\left(\frac{P^g_t/(1-P^g_t)}{Q^g_t/(1-Q^g_t)}\right),
\]
where $P_t^g$ and $Q_t^g$ denote the human-written and AI-rephrased word-use probabilities estimated within country-field group $g$. Thus, the pooled benchmark evaluates all groups against a common word-use baseline, whereas the country-field-specific benchmarks evaluate each group against its own human and AI word-use distributions. 

Using these benchmark distributions, we then estimate the fraction of texts that are likely to be modified by AI using a maximum likelihood estimator (MLE) applied to later-year publications: all publications from 2021 are used in the pre-LLM placebo analysis, and all publications from 2025 are used in the post-LLM AI-use analysis. The method follows \citet{liang2025quantifying}: let $\pi_t^H$ and $\pi_t^A$ denote the human-written and AI-rephrased probabilities of word $t$ under a certain benchmark. For a given benchmark vocabulary $\mathcal{V}$, let $S_i \subseteq \mathcal{V}$ denote the set of benchmark words observed in sentence $i$. We model word occurrence at the sentence level as a Bernoulli event. Thus, under the human-written distribution, the likelihood of sentence $i$ is
\[
L_H(S_i)
=
\prod_{t \in S_i} \pi_t^H
\prod_{t \in \mathcal{V}\setminus S_i} (1-\pi_t^H),
\]
and under the AI-rephrased distribution, it is
\[
L_A(S_i)
=
\prod_{t \in S_i} \pi_t^A
\prod_{t \in \mathcal{V}\setminus S_i} (1-\pi_t^A).
\]
The mixture likelihood for sentence $i$ is therefore
\[
L_i(\alpha)
=
(1-\alpha)L_H(S_i)+\alpha L_A(S_i),
\]
where $\alpha \in [0,1]$ denotes the mixture weight on the AI-rephrased distribution. The normalized log-likelihood is
\[
\ell(\alpha)
=
\frac{1}{N}\sum_{i=1}^{N}
\log\left[
(1-\alpha)+\alpha \exp\left\{
\log L_A(S_i)-\log L_H(S_i)
\right\}
\right].
\]
Finally, we obtain the MLE of $\alpha$ by solving the one-dimensional optimization: $\hat{\alpha}
=
\arg\max_{\alpha\in[0,1]}\ell(\alpha)$.

We apply this estimator in two ways. For the pooled-benchmark estimates, the benchmark probabilities are $(\pi_t^H,\pi_t^A)=(P_t,Q_t)$, where $P_t$ and $Q_t$ are estimated using all sampled publications pooled across country-field groups. For the country-field-specific estimates, the benchmark probabilities are $(\pi_t^H,\pi_t^A)=(P_t^g,Q_t^g)$,
where $P_t^g$ and $Q_t^g$ are estimated using only publications from country-field group $g$. In both cases, we estimate $\alpha_g$ separately within each country-field group: we denote $\alpha_{g}^{pool}$ to be the estimated share of AI-modified text in group $g$ using the \textit{pooled} benchmark, and $\alpha_{g}^{g}$ to the estimated share of AI modified texts in group $g$ using the \textit{group g-specific} benchmark. Comparing these two estimates ($\alpha_{g}^{pool}$ \textit{vs.} $\alpha_{g}^{g}$) allows us to quantify how much the estimated mixture weight changes when the benchmark reflects country-field-specific writing baselines rather than imposing a common pooled baseline across all groups. Additional methodological details are provided in Appendix B.

\subsection*{Pre-LLM Stylistic Differences}
Some scientific writing may have already resembled AI-generated language before the rise of LLMs \cite{rashidi2023chatgpt}. We begin by examining the pre-LLM distribution of “AI-like” words, defined as those with the lowest log-odds ratios, indicating that they are substantially more likely to appear in LLM-rephrased text than in the original human-written abstracts. Because all texts are processed using the same rephrasing procedure, any systematic variations in these log-odds ratios across country-field groups would suggest pre-existing stylistic differences across different country-fields.

Appendix Table A1 reports the top 20 AI-like words that appear in at least 100 country-field groups. For example, ``notably” has the lowest average log-odds ratio close to $-4.1$, implying that it is roughly 60 times more likely to appear in AI-rephrased text than in human-written text. Importantly, there is substantial dispersion: the variance of log-odds ratios ranges from about $0.3$ to $0.67$ across words (column 3). Decomposing this variation using an ANOVA framework, we find that, on average, around 20\% of the total variation is attributable to differences across fields, while the remaining 80\% arises from within-field differences across countries. This pattern indicates that cross-country stylistic heterogeneity within the same discipline is the dominant source of variation.

A natural question is whether these differences simply reflect random noise. Under the null hypothesis that country-field groups do not differ systematically in pre-LLM writing style, variations in log-odds ratios should arise purely from sampling noise. To assess this, we conduct a permutation-based simulation in which country-field labels are randomly reassigned across publications, and the log-odds ratios are recomputed. This procedure preserves the overall distribution of words while eliminating any systematic association between writing style and country-field identity. 

In Appendix Table A1 (columns 6-8), we report simulated results over 50 runs. As expected, the average number of country-field groups in which a word appears increases, reflecting that words previously concentrated in a subset of country-fields are redistributed more evenly across all groups under the simulation. While the average mean log-odds ratios remain similar, the variances are substantially smaller than the actual variances observed in column 3, ranging from 0.03 to 0.27. We conduct F-tests of equality of variances between the actual training sample and each simulated sample with the alternative hypothesis that the simulated sample has a smaller variance than the actual sample. The results show that, for 18 out of the top 20 AI-like words, the simulated variances are significantly smaller with mean p-values below 0.001. These findings indicate that the heterogeneity observed in the empirical data reflects genuine pre-existing stylistic differences rather than statistical noise. In the next section, we examine how such pre-existing differences affect downstream results.

\subsection*{Estimated AI-Like Text Shares With and Without Context}
Interpreting trends and making comparisons across groups requires ex-ante comparability, i.e., a common baseline. We show that ignoring context---specifically, pre-existing stylistic differences across countries and fields---when evaluating AI-likeness in scientific writing undermines this comparability and leads to fragile or biased interpretations. In particular, we apply (i) a pooled benchmark and (ii) 234 country-field-specific benchmarks to all publications in 2021, and compare how the MLE estimated share of AI-modified text differs. 

We estimate separately for each country-field at the quarterly level, and take an average to get the average estimated $\alpha_g$ in 2021. The results are robust to using a finer monthly frequency. To ensure reliable inference, we restrict attention to country-field groups with an average testing sample size above 2,000, matching the size of the sampled training data. The results are consistent but noisier when we lower this cutoff because larger samples yield more precise estimates with MLE variance shrinking as sample size grows.

\begin{figure}[t]
    \centering
    \begin{subfigure}[c]{0.71\linewidth}
        \textbf{a}
        
        \centering
        \includegraphics[width=1\linewidth]{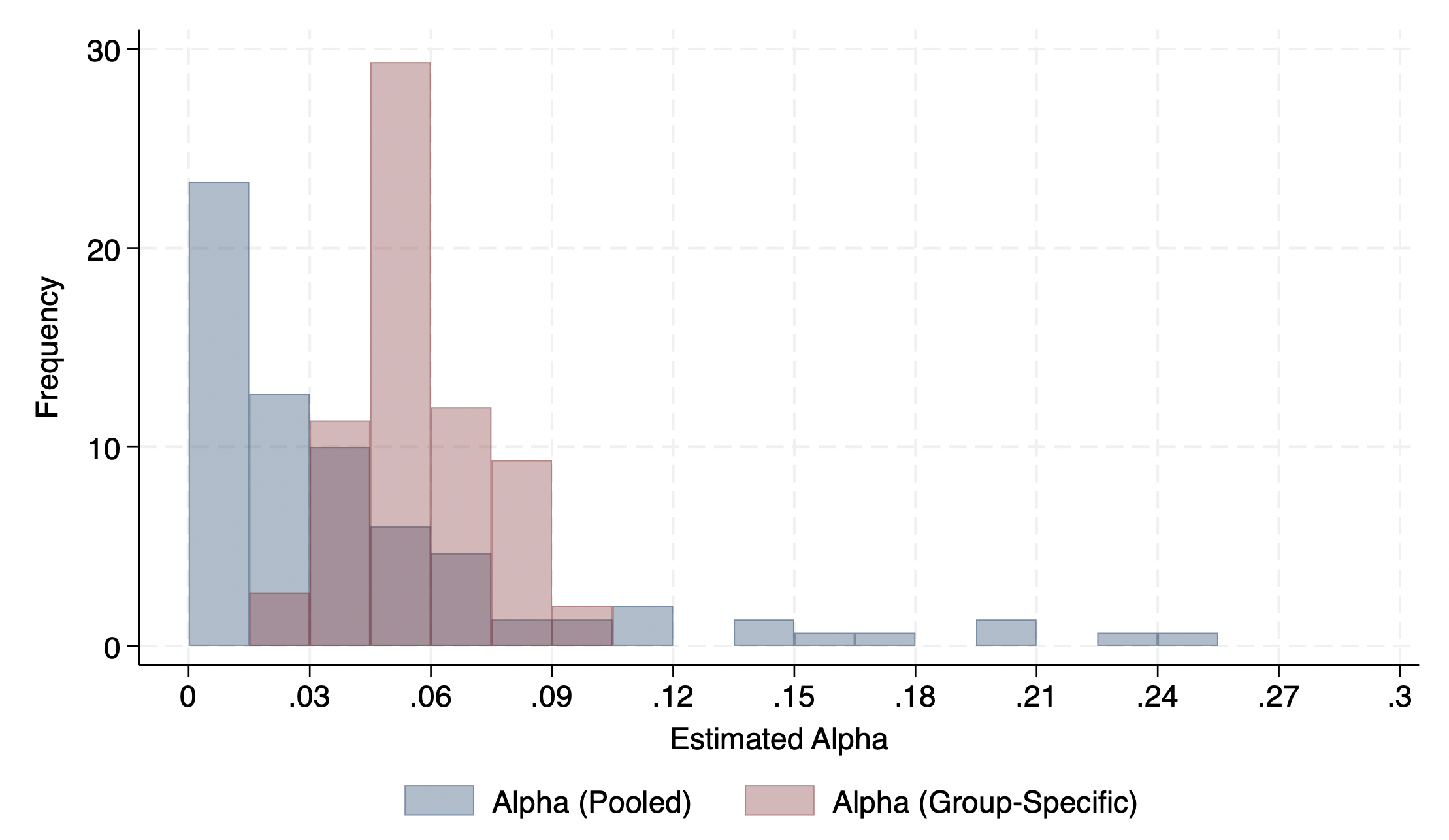}
    \end{subfigure}

    \begin{subfigure}[c]{0.49\linewidth}
        \textbf{b}
        
        \centering
        \includegraphics[width=1\linewidth]{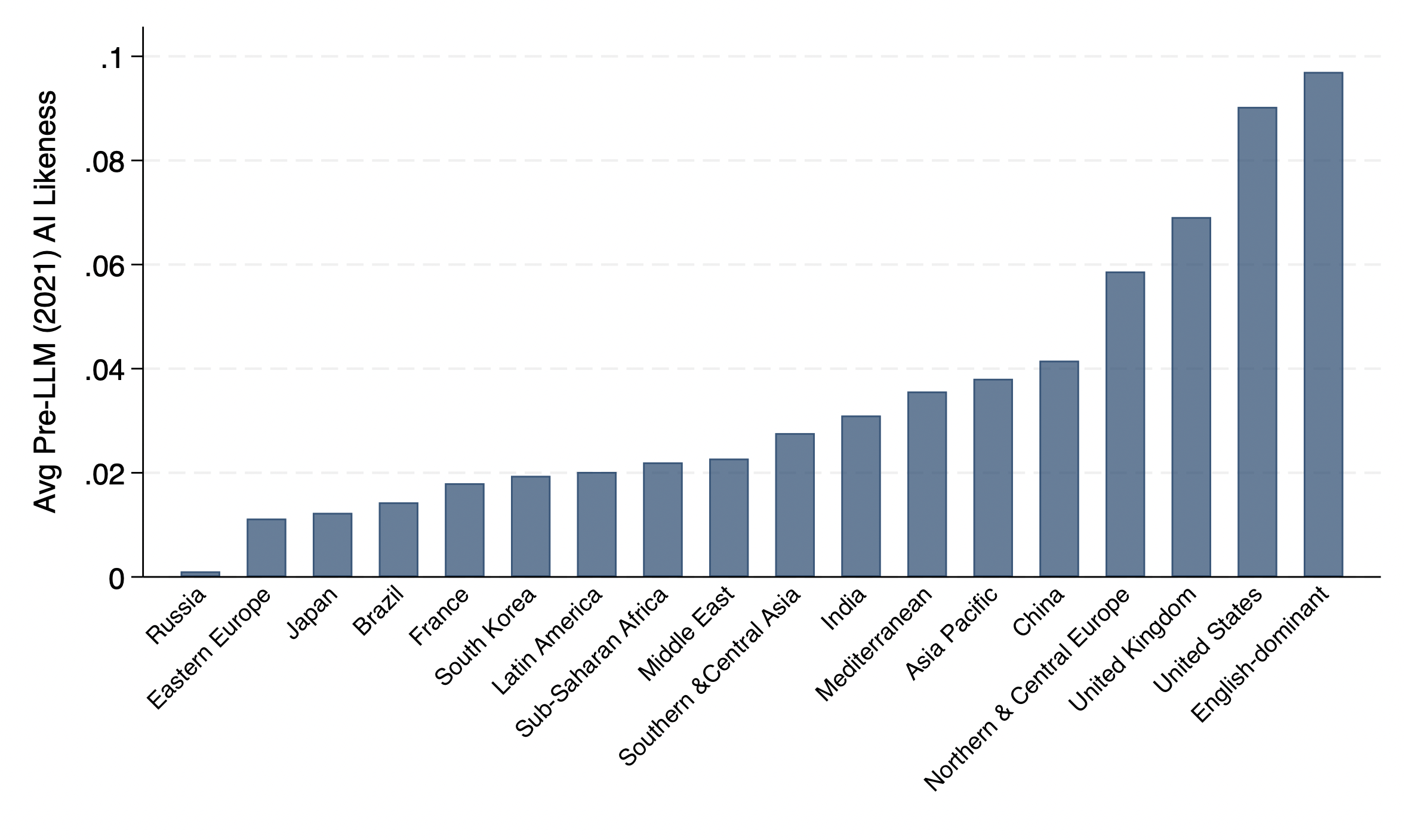}
    \end{subfigure}
    \hfill
    \begin{subfigure}[c]{0.49\linewidth}
    \textbf{c}
    
        \centering
    \raisebox{4pt}{
        \includegraphics[width=\linewidth]{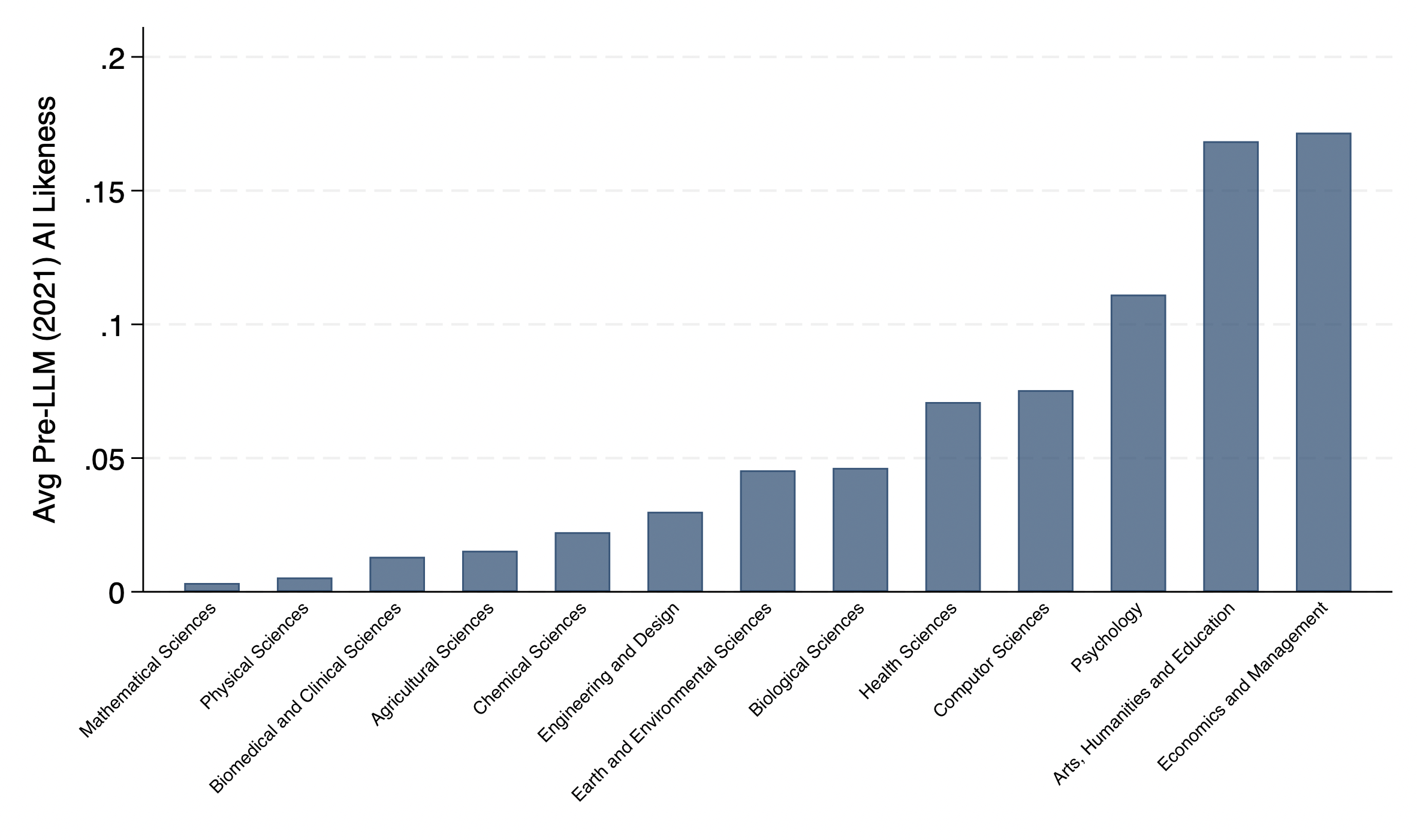}
    }
    \end{subfigure}
    
    \caption{Distribution of Estimated Pre-LLM AI-likeness in 2021}
        \label{fig:alpha_estimation}
    \begin{quote}
    \vspace{0.5\baselineskip}
        \footnotesize{\textit{Note:} This figure shows the distributions of estimated $\alpha$ in 2021. Panel A shows the estimated AI-likeness for country-field groups where the average testing sample size is larger than 2000. The blue bar shows the estimates using a pooled benchmark trained on all sampled publications ($\alpha_g^{pool}$). The red bar shows the estimates using country-field-specific benchmarks, each trained only on publications within the corresponding country-field group ($\alpha_g^{g}$). Panels B and C show the country-level and field-level average estimated AI-likeness in 2021 using the pooled benchmark.}
    \end{quote}
\end{figure}

When applying the pooled AI-likeness benchmark to pre-LLM publications in 2021 across 234 country-field groups, where any AI-like wording should arise purely by chance rather than actual LLM use, we observe substantial variation and a highly skewed distribution in the estimated $\alpha_g^{pool}$. As shown in Figure 1a (blue bars), the estimates have a mean of 0.044 and a standard deviation of 0.052, indicating large cross-group variation despite the absence of true AI usage. Figures 1b and 1c report the corresponding country-level and field-level averages. For example, the average estimated $\alpha^{pool}$ in 2021 is around 0$\sim$0.01 for countries such as Russia, but as high as 0.09 for the United States and other English-dominant countries. Variation across fields is also pronounced: the average estimated $\alpha^{pool}$ is below 0.01 in Mathematical Sciences and Physical Sciences, but rises to 0.17 in Economics and Management and in Arts, Humanities, and Education.

Such dispersion is unlikely to arise from random noise alone. To assess this, we conduct another permutation-based simulation in which country-field labels are randomly reassigned across publications, thereby eliminating any systematic link between writing style and country-field identity. Appendix Figure A1 shows the distribution across 300 simulation rounds: the resulting distributions are approximately normal and centered near 0.035. The contrast between Appendix Figure A1 and the skewed distribution obtained under the pooled benchmark in Figure 1 (blue bars) reflects systematic pre-existing heterogeneity in scientific writing styles across country-fields. 

This skewed baseline distribution undermines comparability and could lead to biased perceptions of AI usage across countries and fields. In pre-LLM data, $\alpha$ should not be interpreted as actual AI use; rather, it captures the share of wording that overlaps, by chance, with the benchmark's AI-like word profile. A very high baseline $\alpha$ is therefore problematic because it implies that some groups are mechanically closer to the pooled AI-like profile before the rise of LLMs. At the same time, a baseline of exactly zero would also be implausible, since human-written scientific text can naturally contain words that later become characteristic of LLM-rephrased writing. A credible benchmark should therefore produce a small but nonzero average pre-LLM $\alpha$, with limited dispersion across groups. 

To achieve this, we propose to incorporate context into the benchmark by constructing country-field-specific benchmarks, each trained only on publications within the corresponding country-field group. Under this contextualized approach, the pre-LLM variation in estimated $\alpha_g^g$ across country-fields is substantially attenuated (red bars in Figure 1a): the dispersion is greatly reduced (SD $=$ 0.016), and the mean remains similar, indicating that roughly 4$\sim$5\% of observed wording may overlap by chance with words later identified as relatively AI-like by the MLE benchmark. The resulting distribution is closer to normal, mitigating distortions induced by the single pooled benchmark and providing a more credible baseline for cross-group comparison.

\subsection*{De-biasing AI-Likeness Estimates}
Relative to our country-field-specific benchmarks, we examine how AI usage after the rise of LLMs may be mismeasured when using a pooled benchmark that ignores context. We do not claim that our contextual benchmarks are fully unbiased; rather, our goal is to demonstrate the existence of bias and suggest the direction of bias with the new approach.

Specifically, we estimate AI usage for all publications in 2025 using our country-field-specific benchmarks ($\alpha_{g}^{g}$), and compare these estimates with estimates obtained using the pooled benchmark ($\alpha_{g}^{pool}$). For each country-field group, we compute the demeaned log ratio of the two estimates. Positive values indicate that AI usage is underestimated under the pooled benchmark ($\alpha_{g}^{g} > \alpha_{g}^{pool}$), while negative values indicate overestimation ($\alpha_{g}^{g} < \alpha_{g}^{pool}$), relative to the contextual benchmark. Figure 2 reports the average demeaned log ratio aggregated by country (Panel A) and by field (Panel B).

\begin{figure}[htbp]
    \centering
    \begin{subfigure}[c]{0.49\linewidth}
        \textbf{a}
        
        \centering
        \includegraphics[width=1\linewidth]{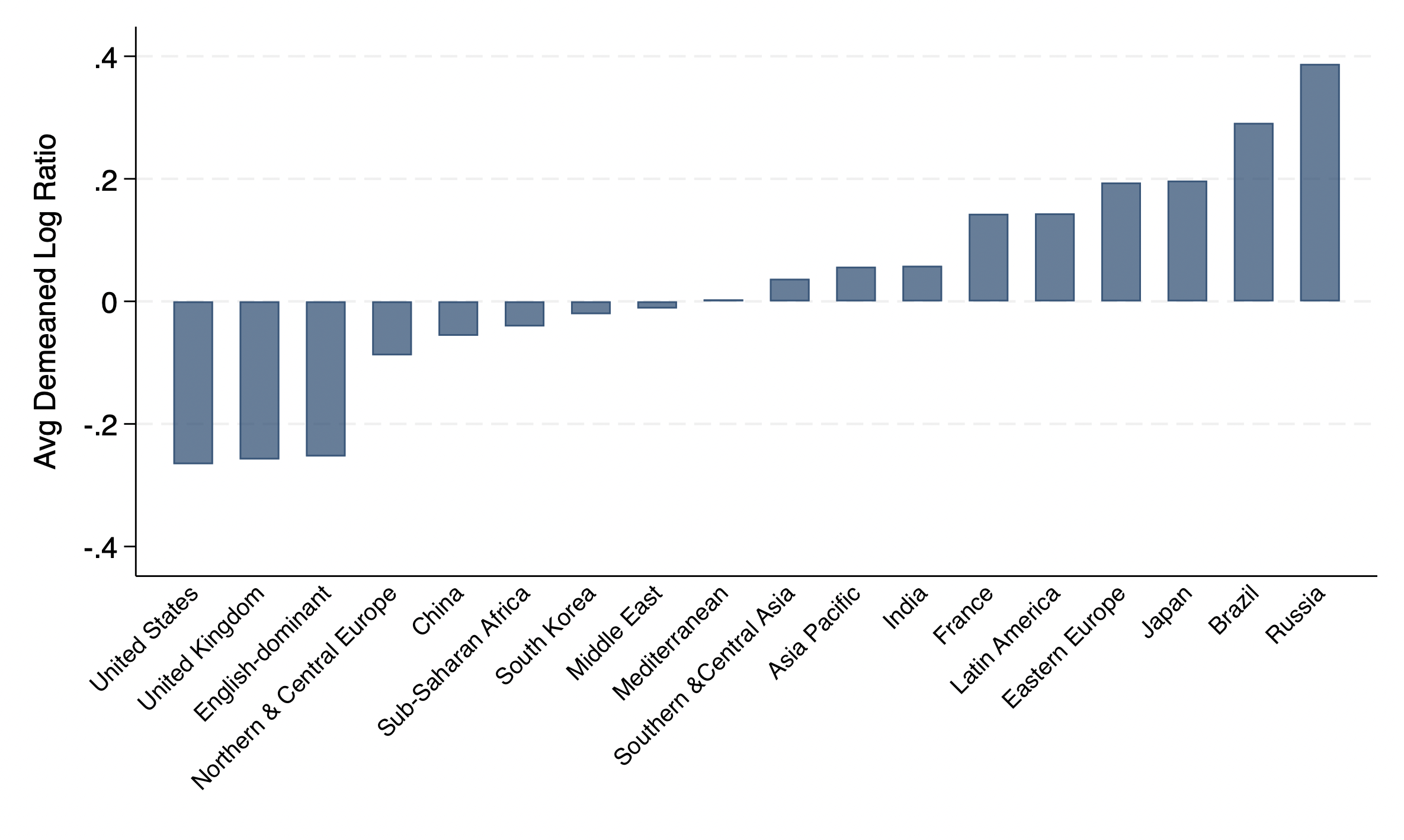}
    \end{subfigure}
    \hfill
    \begin{subfigure}[c]{0.49\linewidth}
    \textbf{b}
    
        \centering
    \raisebox{4pt}{
        \includegraphics[width=\linewidth]{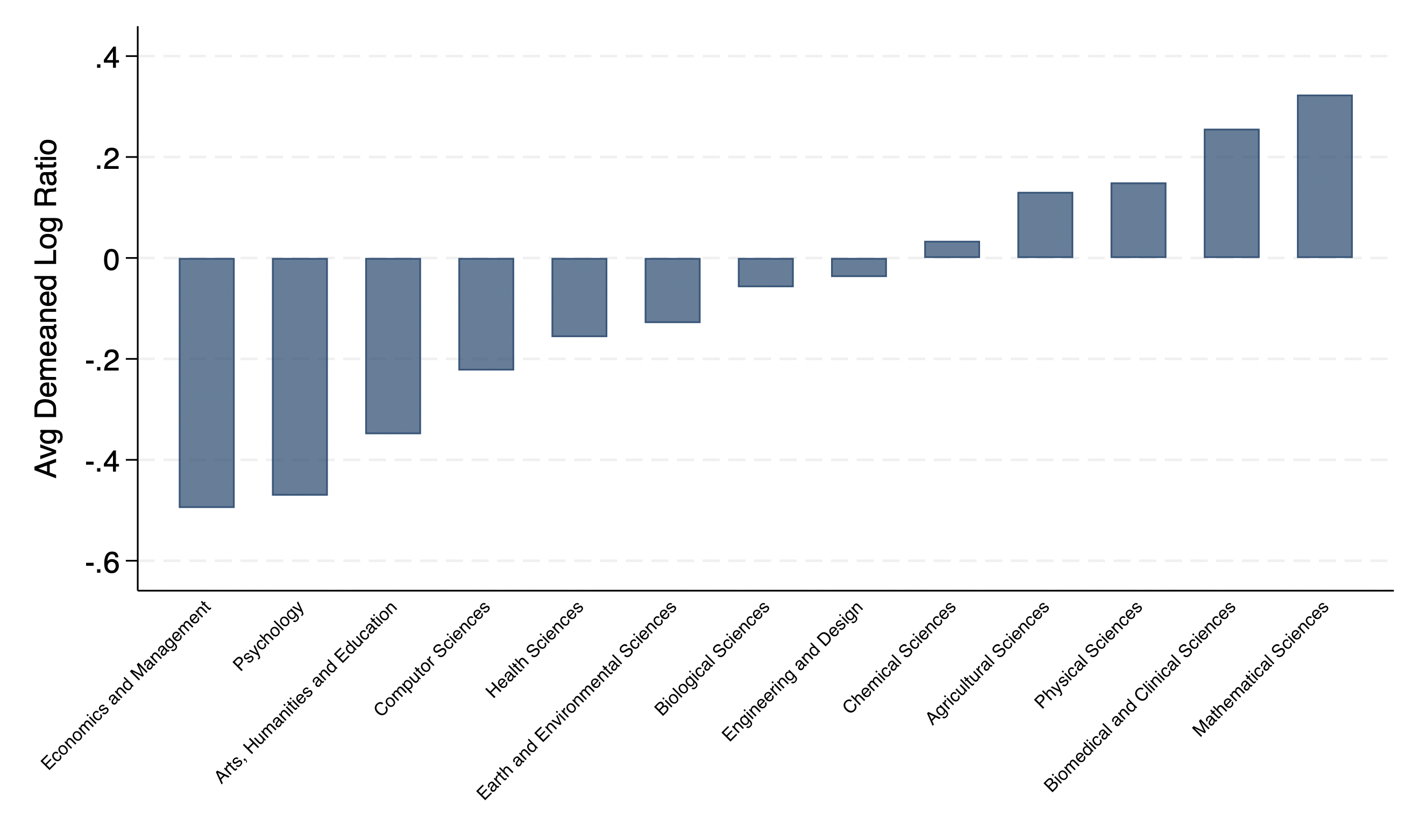}
    }
    \end{subfigure}
    \caption{Biased Perceptions in Academic Writing AI Evaluation}
    \vspace{0.5\baselineskip}
    \begin{quote}
        \footnotesize{\textit{Note:} This figure reports the average log ratio for each country (Panel A) and each field (Panel B). The log ratio is computed as the demeaned logarithm of the ratio between the pooled estimates and the corresponding country-field-specific estimates of $\alpha$ in 2025. Positive values indicate that AI usage in writing is underestimated under the pooled benchmark, while negative values indicate overestimation. }
    \end{quote}
\end{figure}

The results reveal systematic biases. Geographically, AI usage is most strongly overestimated in the United States, the United Kingdom, and other English-dominant countries, with similar but smaller biases in Northern \& Central Europe and China. In contrast, AI usage is underestimated in countries such as Russia, Brazil, Japan, and Eastern Europe. These patterns align closely with pre-existing stylistic differences: groups whose baseline writing is closer to the ``AI-like” profile tend to be mechanically assigned higher AI usage.

A similar pattern emerges across disciplines. Fields characterized by baseline writing that more closely resembles the AI-like profile---such as Economics and Management, Psychology, Arts, Humanities and Education, and Computer Science---exhibit pronounced overestimation under the pooled benchmark. In contrast, fields whose baseline writing is less aligned with the pooled AI-like profile, including Mathematical Sciences, Biomedical Sciences, Physical Sciences, and Agricultural Sciences, show systematic underestimation.

Taken together, these results demonstrate that a single pooled benchmark does not merely introduce noise; it induces structured bias that reshapes the perceived geographic and disciplinary distribution of AI usage in scientific writing. Incorporating context, therefore, helps mitigate these distortions and makes the estimates more credible.

\subsection*{Discussion and Policy Implications}
This paper shows that evaluating AI usage in scientific writing without accounting for context can lead to systematic bias in the perceived prevalence of AI use. In particular, pooled benchmarks may confound true AI usage with pre-existing stylistic differences across countries and fields, generating distorted comparisons even in pre-LLM settings where no AI is used. Our country-field-specific benchmarks incorporating these contexts can meaningfully reduce the distortions and improve the credibility of cross-group comparisons.

These findings have several important implications. First, they reshape how we interpret who is ``using AI.” Our results suggest that the patterns we perceive may largely reflect stylistic alignment with the benchmark rather than true differences in AI use. As a result, public narratives about AI usage in science---whether in media, policy discussions, or academic discourse---may become distorted, reinforcing inaccurate stereotypes about regions or disciplines. Policymakers and analysts should therefore exercise caution when interpreting cross-group differences in estimated AI usage and avoid drawing conclusions without accounting for contextual baselines.

Second, these distortions can contribute to an erosion of trust that is unevenly distributed across groups. If certain countries or disciplines are perceived to overuse AI, readers may discount the credibility of their work \cite{raj2026artificial}, even when such perceptions are driven by bias rather than actual behavior. This concern might be salient in certain disciplines: for example, people may overestimate the use of LLM in Economics and Management or the humanities, where the baseline writing style more closely resembles the AI-like profile, which could lead to unwarranted skepticism toward social science research and narratives from the public.

The consequences may be even more pronounced across countries. While prior work suggests that AI tools can improve equity in science by assisting non-native English speakers \citep{berdejo2023ai}, biased detection or evaluation frameworks may introduce a new layer of inequality. This is especially important when the consequences of mismeasurement are likely to be asymmetric: for researchers in English-dominant, high-income countries such as the United States, ``AI-like” writing may be interpreted as a signal of fluency or professionalism, whereas for researchers in countries such as China, the same signal may more readily trigger doubts about originality, effort, or credibility. Increasing awareness of these asymmetries is critical so that evaluators, editors, reviewers, and policymakers can better contextualize their judgments.

Third, a biased benchmark design may unintentionally incentivize convergence in writing style, reducing diversity in scientific expression. If researchers anticipate that certain styles are more likely to be flagged as ``AI-like,” they may adapt their writing to align with perceived norms, potentially homogenizing scientific communication. This concern extends beyond stylistic diversity. Emerging evidence suggests that biased AI-assisted writing tools can subtly influence users’ attitudes without explicit awareness \citep{williams2026biased}. In this sense, biases in AI evaluation frameworks may not only distort measurement but may also shape the evolution of scientific discourse itself. Preserving diversity in writing styles is therefore important not only for fairness but also for maintaining intellectual plurality in science.

Finally, our results highlight the risks of one-size-fits-all AI detection and evaluation standards. Policies that rely on pooled benchmarks---especially those trained predominantly on Western or English-dominant contexts---may introduce systematic bias when applied globally. This concern is consistent with evidence that existing AI detection models (e.g., GPT detectors) can exhibit group-level biases \citep{liang2023gpt}. As a result, this has direct implications for academic integrity policies, journal screening practices, funding compliance checks, and broader monitoring of AI usage. Rather than adopting uniform thresholds or models, decision-makers should consider context-calibrated benchmarks or, at a minimum, be aware of the potential bias and account for baseline differences across countries and fields in their decision-making.

Taken together, these findings suggest that accurately evaluating AI usage in scientific writing requires moving beyond pooled approaches toward context-aware measurement. While no benchmark is perfectly unbiased, incorporating contextual information represents an important step toward more reliable, equitable, and interpretable assessments. In a scientific system already shaped by uneven evaluative standards, measurement tools should not introduce additional sources of bias.

\newpage
\setstretch{1}

\renewcommand{\refname}{\center{REFERENCES}}
\bibliography{reference}

\newpage
\setstretch{1.5}

\appendix
\setcounter{table}{0}\renewcommand{\thetable}{A\arabic{table}}
\setcounter{figure}{0}\renewcommand{\thefigure}{A\arabic{figure}}
\section{Additional Tables and Figures}

\begin{table}[htbp]
\centering
\caption{Country-field Heterogeneity in Writing: Top AI-like Words}
\label{tab:variance_comparison}
\scriptsize
\setlength{\tabcolsep}{3pt}
\resizebox{\textwidth}{!}{
\begin{tabular}{lccccc|cccl}
\toprule
& \multicolumn{5}{c}{\textbf{Training Sample}} & \multicolumn{3}{c}{\textbf{Simulation}} \\
\cmidrule(lr){2-6} \cmidrule(lr){7-9}
Word & N & Mean & Variance & Cross & Within 
& Average & Average  & Average & Average\\
&&&  & Field \% & Field \% & N & Mean & Variance & P-value\\
& (1) & (2) & (3) & (4) & (5) & (6) & (7) & (8) & (9)\\
\midrule
notably      & 219 & -4.081 & 0.404 & 0.223 & 0.777 & 231 & -3.971 & 0.101 & 0.000$^{****}$ \\
alongside    & 190 & -3.788 & 0.666 & 0.276 & 0.724 & 224 & -3.644 & 0.174 & 0.000$^{****}$ \\
notable      & 213 & -3.700 & 0.303 & 0.239 & 0.761 & 223 & -3.718 & 0.189 & 0.004$^{***}$ \\
utilizing    & 224 & -3.155 & 0.598 & 0.107 & 0.893 & 233 & -2.916 & 0.079 & 0.000$^{****}$ \\
advancements & 127 & -2.988 & 0.376 & 0.168 & 0.832 & 165 & -3.023 & 0.228 & 0.003$^{***}$ \\
ultimately   & 225 & -2.916 & 0.284 & 0.098 & 0.902 & 231 & -2.829 & 0.109 & 0.000$^{****}$ \\
emphasizing  & 155 & -2.897 & 0.319 & 0.102 & 0.898 & 179 & -2.977 & 0.242 & 0.063$^{*}$ \\
intricate    & 116 & -2.848 & 0.496 & 0.217 & 0.783 & 141 & -2.985 & 0.226 & 0.000$^{****}$ \\
employs      & 174 & -2.841 & 0.411 & 0.237 & 0.763 & 223 & -2.775 & 0.179 & 0.000$^{****}$ \\
outlines     & 118 & -2.781 & 0.493 & 0.381 & 0.619 & 169 & -2.927 & 0.250 & 0.000$^{****}$ \\
additionally & 234 & -2.713 & 0.319 & 0.273 & 0.727 & 234 & -2.599 & 0.027 & 0.000$^{****}$ \\
specifically & 234 & -2.684 & 0.496 & 0.440 & 0.560 & 234 & -2.512 & 0.026 & 0.000$^{****}$ \\
serves       & 220 & -2.683 & 0.338 & 0.210 & 0.790 & 227 & -2.610 & 0.169 & 0.000$^{****}$ \\
emphasizes   & 178 & -2.640 & 0.361 & 0.278 & 0.722 & 207 & -2.709 & 0.240 & 0.010$^{**}$ \\
conversely   & 212 & -2.603 & 0.414 & 0.150 & 0.850 & 228 & -2.500 & 0.157 & 0.000$^{****}$ \\
highlighting & 205 & -2.582 & 0.525 & 0.123 & 0.877 & 231 & -2.434 & 0.121 & 0.000$^{****}$ \\
thorough     & 198 & -2.499 & 0.328 & 0.098 & 0.902 & 219 & -2.523 & 0.215 & 0.007$^{***}$ \\
introduces   & 202 & -2.483 & 0.427 & 0.454 & 0.546 & 230 & -2.490 & 0.121 & 0.000$^{****}$ \\
fostering    & 111 & -2.481 & 0.497 & 0.295 & 0.705 & 173 & -2.510 & 0.274 & 0.001$^{****}$ \\
necessity    & 219 & -2.471 & 0.528 & 0.258 & 0.742 & 225 & -2.465 & 0.178 & 0.000$^{****}$ \\
\bottomrule
\end{tabular}
}
    \begin{quote}
    \vspace{1\baselineskip}
        \footnotesize{\textit{Note:} This table reports summary statistics of log-odds ratios for the top 20 AI-like words, restricting to words that appear in at least 100 of the 234 country-field groups. Coverage varies because some words are infrequently used in certain country-fields, making their odds ratios unreliable or undefined. Columns 1-5 present statistics from the actual training sample, including the mean log-odds ratio, variance, and an ANOVA decomposition of variance into between-field and within-field (across countries) components. More negative log-odds ratios indicate words that are more likely to exist in LLM-modified texts. Columns 6-8 report corresponding average statistics from 50 simulations in which publications are randomly reassigned to country-field groups. Column 9 reports the average p-value from F-tests comparing variances in the training sample and the simulated data. *, **, ***, and **** indicate significance at the 10\%, 5\%, 1\%, and 0.1\% levels.}
    \end{quote}
\end{table}

\newpage
\FloatBarrier
\begin{figure}[htbp]
    \centering
    \includegraphics[width=0.8\linewidth]{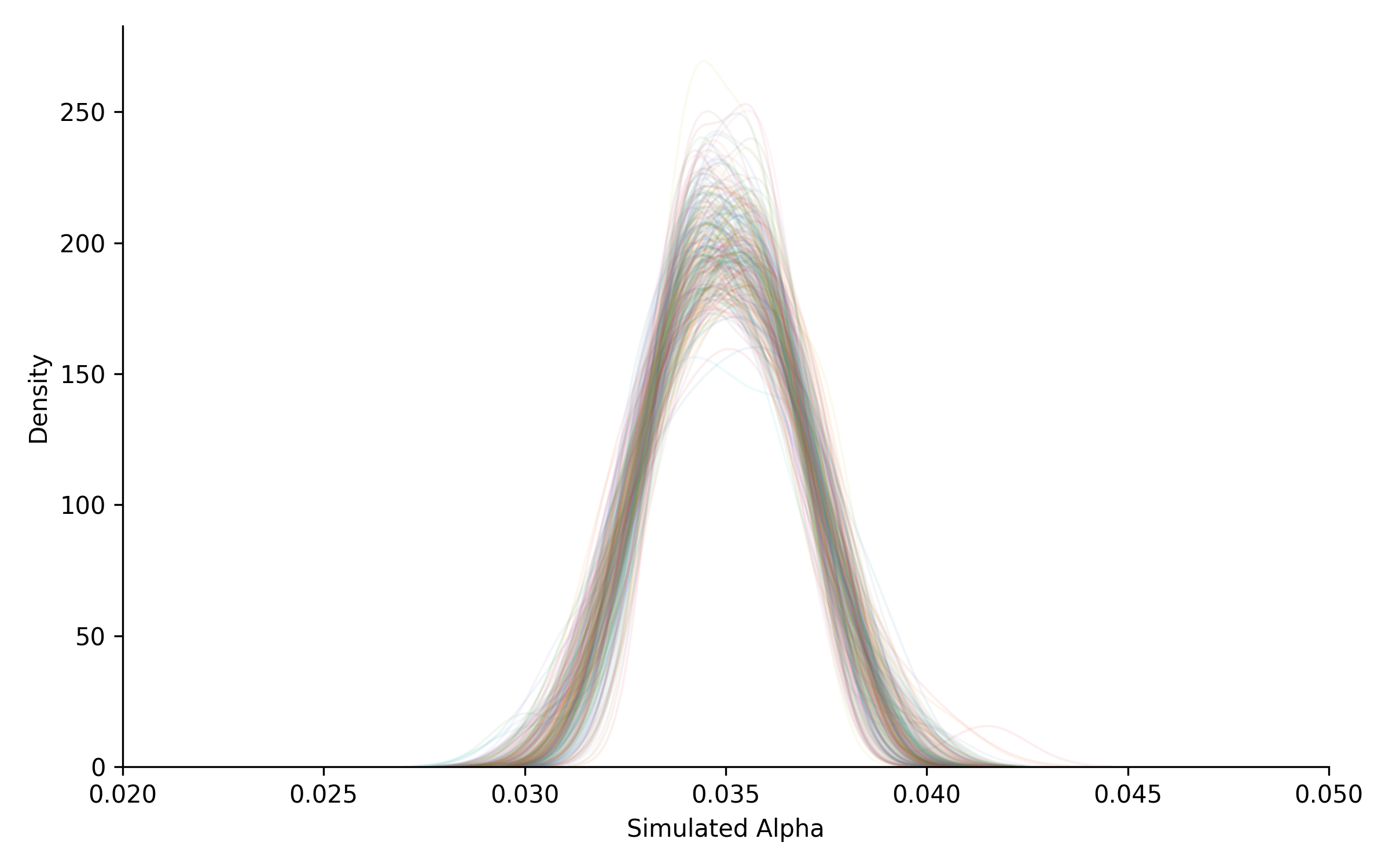}
    \caption{Simulated Distribution of Pre-LLM AI-Likeness}
    \vspace{0.5\baselineskip}
    \begin{quote}
        \footnotesize{\textit{Note:} This figure shows the distribution of average estimated $\alpha$ across country-field groups in 300 rounds of permutation-based simulation where country-field labels are randomly reassigned across publications.}
    \end{quote}
    \label{fig:sim_2}
\end{figure}

\FloatBarrier

\newpage
\setcounter{table}{0}\renewcommand{\thetable}{B\arabic{table}}
\setcounter{figure}{0}\renewcommand{\thefigure}{B\arabic{figure}}
\section{Data and Methods}

\subsection{Publication Sampling}
\label{app:sampling}
We use the Dimensions database\footnote{For more information about Dimensions, visit https://www.dimensions.ai}, accessed through the MIT institutional subscription. Dimensions aggregates digitized scholarly publications from public metadata sources (e.g., Crossref, PubMed, arXiv) and publisher feeds. Owing to its scale, broad coverage, and rich metadata, it is often used in bibliometric and science-of-science research. From this database, we extract all English-language publications from 2020 to 2025.

In 2020, the dataset contains approximately 2.2 million publications, of which about 1.8 million include at least one author with identifiable affiliation information that maps to a country. We assign publications to countries using fractional counting based on author affiliations. For example, a paper with three authors---two affiliated with the United States and one with China---is assigned $\frac{2}{3}$ to the United States and $\frac{1}{3}$ to China. Our results are robust to an alternative sampling restriction that focuses only on publications whose authors are all affiliated with the same country group.

\begin{table}[htbp]
\centering
\caption{Country Group Classification}
\begin{tabular}{l l l}
\hline
\textbf{Country / Group} & \textbf{Example Language(s)} & \textbf{Example Countries} \\
\hline
\multicolumn{3}{l}{\textit{Single Countries}} \\
\hline
Brazil & Portuguese & --- \\
China & Mandarin Chinese & --- \\
France & French & --- \\
India & Hindi, English & --- \\
Japan & Japanese & --- \\
Russia & Russian & --- \\
South Korea & Korean & --- \\
United Kingdom & English & --- \\
United States & English & --- \\
\hline
\multicolumn{3}{l}{\textit{Country Groups}} \\
\hline
Asia Pacific & Sino-Tibetan, Austronesian & Singapore, Malaysia, Philippines \\
Eastern Europe & Slavic, Romance, Albanian & Poland, Romania, Serbia \\
English-dominant & English & Australia, Canada, New Zealand \\
Latin America \& Caribbean & Spanish, Portuguese & Mexico, Brazil, Argentina \\
Middle East & Arabic, Hebrew, Turkish & Saudi Arabia, Israel \\
Northern \& Central Europe & Germanic, Uralic, Slavic & Germany, Sweden, Poland \\
Southern \& Central Asia & Indo-Aryan, Dravidian & India, Pakistan, Bangladesh \\
Mediterranean & Romance, Greek & Italy, Spain, Greece \\
Sub-Saharan Africa & Niger-Congo, Afroasiatic & South Africa, Kenya, Nigeria \\
\hline
\end{tabular}
\end{table}

In total, 198 countries and regions are represented in the 2020 data. We aggregate these into 18 country groups based on linguistic and geographic proximity, as detailed in Appendix Table B1.

Dimensions classifies publications into 22 fields of study based on the 2020 Australian and New Zealand Standard Research Classification (ANZSRC). We further aggregate these into 13 broader fields based on disciplinary similarity and field size as shown in Appendix Figure B2. In particular, smaller fields, especially in the social sciences and humanities (e.g., History, Philosophy), are combined to ensure sufficient sample sizes for analysis.

\begin{table}[htbp]
\centering
\caption{Field of Study Classification}
\begin{tabular}{l c}
\hline
\textbf{Field Group} & \textbf{Number of Publications} \\
\hline
Biomedical and Clinical Sciences & 447{,}772 \\
Engineering and Design & 384{,}234 \\
Chemical Sciences & 173{,}745 \\
Biological Sciences & 159{,}796 \\
Physical Sciences & 121{,}604 \\
Earth and Environmental Sciences & 106{,}214 \\
Health Sciences & 91{,}852 \\
Agricultural Sciences & 70{,}435 \\
Computer Sciences & 70{,}344 \\
Arts, Humanities, and Education & 65{,}784 \\
Mathematical Sciences & 58{,}387 \\
Economics and Management & 44{,}860 \\
Psychology & 41{,}388 \\
\hline
\end{tabular}
\end{table}

The 18 country groups and 13 field groups yield 234 country-field groups. For each group, we randomly sample up to 2,000 publications. We follow two schemes to assign publications to countries: (i) fractional allocation based on author affiliations, and (ii) restricting to publications whose authors are all from the same country. Our results are robust to the choice of scheme. For country-field groups with fewer than 2,000 publications in 2020, we include all available publications.

\newpage
\subsection{AI Rewriting Procedure}
We collect the abstracts for all sampled publications. The final training dataset contains 492,086 unique human-written abstracts. For each abstract, we generate a corresponding AI-written version using \texttt{gpt-4o-mini-2024-07-18}. Our prompt follows the outline-and-expand structure of \citet{liang2025quantifying}, but compresses the procedure into a one-turn interaction: the model first summarizes the abstract into an outline, then expands it into a rewritten abstract of similar length, as illustrated in Appendix Figure B1.

\begin{figure}[!ht]
\centering
\small
\begin{tcolorbox}[fonttitle=\fontfamily{pbk}\selectfont\bfseries,
                  fontupper=\fontsize{9}{9}\fontfamily{ppl},
                  fontlower=\fontfamily{put}\selectfont\scshape,
                  title=Evaluation Prompt for the Preference Comparison Experiment,
                  width=\linewidth,
                  arc=1mm, auto outer arc]
\begin{Verbatim}[breaklines=true, breaksymbol={}]
{
    "content": "You are a helpful and conversational AI assistant. Respond to the following task.",
    "role": "system"
},
{
    "content": "You have experience in academic writing. Your task is to paraphrase a given academic text without altering its meaning, using a structured two-step process in a single response:

        1. **Outline Creation**
        - Reverse-engineer the author's writing process by **taking a piece of text** from a paper and **compressing it into a concise outline**.
        - Begin by briefly identifying the *goal or section type* of the text (e.g., introduction, methods, results, discussion).
        - Then, **produce a short set of bullet points** summarizing the core ideas, arguments, or findings.
        - This should reflect how an author might distill their thoughts into a structured plan.
            
        2. **Expansion**
        - Your objective is to **expand upon the** concise version **previously crafted**. 
        - Use the bullet points to write a **detailed, structured narrative**.
        - The expanded text should be roughly similar in length to the original text (around {len(row['Abstract'].split())} words).
        
        After completing the outline and expanded narrative, validate briefly that each step in your checklist has been fulfilled and the meaning of the original text is preserved; if not, self-correct minimally.

        Input text to process:
        \"\"\"{row['Abstract']}\"\"\",
    "role": "user"
},

\end{Verbatim}
\end{tcolorbox}
\caption{Prompt used to obtain AI-generated abstracts for human abstracts} 
\label{fig:prompt}
\end{figure}

\subsection{Log-Odds Ratio Construction}
\label{app:log_ratio}
For the pooled benchmark, we ignore country-field labels and combine all sampled 2020 publications into one training corpus. For each publication $i$, let $H_i$ denote the human-written abstract, $A_i$ the AI-rewritten abstract, and $w_i$ the publication fraction weight, constructed from author affiliations as described in Appendix \ref{app:sampling}. Under fractional counting, a publication contributes only the fraction assigned to the sampled country-field group. For example, if a publication is split equally between the United States and the United Kingdom but enters the sample only through the U.S.-field group, it receives weight $w_i=0.5$ in both the pooled benchmark and the corresponding country-field-specific benchmark. After lowercasing and tokenizing abstracts into alphabetic words, we compute weighted word counts
\[
C^H_t = \sum_i w_i n_t(H_i),
\qquad
C^A_t = \sum_i w_i n_t(A_i),
\]
where $n_t(\cdot)$ is the count of word $t$ in an abstract. We retain words with at least five weighted occurrences in both corpora and define the pooled vocabulary as their intersection, $\mathcal{V}$. For each $t \in \mathcal{V}$, we estimate
\[
P_t = \frac{C^H_t}{\sum_{s \in \mathcal{V}} C^H_s},
\qquad
Q_t = \frac{C^A_t}{\sum_{s \in \mathcal{V}} C^A_s},
\]
where $P_t$ and $Q_t$ are the human and AI word-use probabilities, respectively. The pooled log-odds ratio is
\[
\text{log-odds ratio}_t
=
\log\left(\frac{P_t/(1-P_t)}{Q_t/(1-Q_t)}\right).
\]
More negative values indicate words used relatively more often in AI-rewritten abstracts, while more positive values indicate words used relatively more often in human-written abstracts.

For the country-field-specific benchmarks, we apply the same procedure separately within each country-field group $g$. Specifically, we compute
\[
C^{H,g}_t = \sum_{i \in g} w_i n_t(H_i),
\qquad
C^{A,g}_t = \sum_{i \in g} w_i n_t(A_i),
\]
restrict the vocabulary to words with at least one weighted occurrence in both corpora within $g$, and estimate
\[
P^g_t = \frac{C^{H,g}_t}{\sum_{s \in \mathcal{V}_g} C^{H,g}_s},
\qquad
Q^g_t = \frac{C^{A,g}_t}{\sum_{s \in \mathcal{V}_g} C^{A,g}_s}.
\]
The country-field-specific log-odds ratio is
\[
\text{log-odds ratio}^g_t
=
\log\left(\frac{P^g_t/(1-P^g_t)}{Q^g_t/(1-Q^g_t)}\right).
\]
Thus, the pooled benchmark evaluates all groups against a common word-use baseline, whereas the country-field-specific benchmark evaluates each group against its own human and AI word-use distributions.

\subsection{Maximum Likelihood Estimation of the AI-Rephrased Mixture Weight}
\label{app:mle}

Here we describe how we estimate the mixture weight on the AI-rephrased benchmark distribution. The procedure follows \citet{liang2025quantifying}, adapted to compare estimates obtained using either the pooled benchmark or the country-field-specific benchmarks described in \ref{app:log_ratio}.

When estimating $\alpha$ for a given set of observed publications, we first choose the benchmark to be used: either the pooled benchmark or the corresponding country-field-specific benchmark. We denote the vocabulary of this selected benchmark by $\mathcal{V}$. Let $\pi_t^H$ and $\pi_t^A$ denote the human-written and AI-rephrased probabilities of word $t$ under that benchmark. For the pooled benchmark,
\[
(\pi_t^H,\pi_t^A)=(P_t,Q_t),
\]
where $P_t$ and $Q_t$ are estimated from the pooled 2020 training corpus. For the country-field-specific benchmark for group $g$,
\[
(\pi_t^H,\pi_t^A)=(P_t^g,Q_t^g),
\]
where $P_t^g$ and $Q_t^g$ are estimated using only the 2020 training corpus from country-field group $g$.

The inference data consist of tokenized abstracts from observed publications. We split each abstract into sentences and treat each sentence as one observation. For sentence $i$, let $S_i=\{t\in\mathcal{V}: t \text{ appears in sentence } i\}$
denote the set of benchmark-vocabulary words appearing in the sentence. Words not included in the benchmark vocabulary are excluded from $S_i$, and sentences with fewer than two tokens are excluded.

We model word occurrence at the sentence level as a Bernoulli event over the benchmark vocabulary. Under the human-written benchmark distribution, the likelihood of sentence $i$ is
\[
L_H(S_i)
=
\prod_{t\in S_i}\pi_t^H
\prod_{t\in\mathcal{V}\setminus S_i}(1-\pi_t^H).
\]
Under the AI-rephrased benchmark distribution, the likelihood is
\[
L_A(S_i)
=
\prod_{t\in S_i}\pi_t^A
\prod_{t\in\mathcal{V}\setminus S_i}(1-\pi_t^A).
\]
Equivalently, the corresponding log-likelihoods are
\[
\log L_H(S_i)
=
\sum_{t\in S_i}\log \pi_t^H
+
\sum_{t\in\mathcal{V}\setminus S_i}\log(1-\pi_t^H),
\]
and
\[
\log L_A(S_i)
=
\sum_{t\in S_i}\log \pi_t^A
+
\sum_{t\in\mathcal{V}\setminus S_i}\log(1-\pi_t^A).
\]

For computational efficiency, we precompute
\[
C_H=\sum_{t\in\mathcal{V}}\log(1-\pi_t^H),
\qquad
C_A=\sum_{t\in\mathcal{V}}\log(1-\pi_t^A).
\]
Then, for each sentence,
\[
\log L_H(S_i)
=
\sum_{t\in S_i}\log \pi_t^H
+
C_H
-
\sum_{t\in S_i}\log(1-\pi_t^H),
\]
and
\[
\log L_A(S_i)
=
\sum_{t\in S_i}\log \pi_t^A
+
C_A
-
\sum_{t\in S_i}\log(1-\pi_t^A).
\]

We assume each observed sentence is generated from a mixture of the human-written and AI-rephrased benchmark distributions. The mixture likelihood for sentence $i$ is
\[
L_i(\alpha)
=
(1-\alpha)L_H(S_i)+\alpha L_A(S_i),
\]
where $\alpha\in[0,1]$ denotes the mixture weight on the AI-rephrased benchmark distribution.

For $N$ observed sentences, the sample log-likelihood is
\[
\tilde{\ell}(\alpha)
=
\frac{1}{N}\sum_{i=1}^{N}
\log\left[
(1-\alpha)L_H(S_i)+\alpha L_A(S_i)
\right].
\]
Since $L_H(S_i)$ is constant with respect to $\alpha$, dropping it from the log-likelihood does not change the maximizing value of $\alpha$. Thus, we maximize the normalized log-likelihood:
\[
\tilde{\ell}(\alpha)
=
\frac{1}{N}\sum_{i=1}^{N}
\log L_H(S_i)
+
\frac{1}{N}\sum_{i=1}^{N}
\log\left[
(1-\alpha)
+
\alpha
\frac{L_A(S_i)}{L_H(S_i)}
\right].
\]
The first term does not depend on $\alpha$, so maximizing $\tilde{\ell}(\alpha)$ is equivalent to maximizing the objective
\[
\ell(\alpha)
=
\frac{1}{N}\sum_{i=1}^{N}
\log\left[
(1-\alpha)
+
\alpha
\frac{L_A(S_i)}{L_H(S_i)}
\right].
\]
Equivalently, 
$
\ell(\alpha)
=
\frac{1}{N}\sum_{i=1}^{N}
\log\left[
(1-\alpha)
+
\alpha
\exp\left\{
\log L_A(S_i)-\log L_H(S_i)
\right\}
\right]
$.
The maximum likelihood estimate is $\hat{\alpha}^{MLE}
=
\arg\max_{\alpha\in[0,1]}\ell(\alpha)$.
In implementation, we estimate $\alpha$ by minimizing the negative normalized log-likelihood,
\[
-\ell(\alpha)
=
-\frac{1}{N}\sum_{i=1}^{N}
\log\left[
(1-\alpha)
+
\alpha
\exp\left\{
\log L_A(S_i)-\log L_H(S_i)
\right\}
\right],
\]
using a bounded numerical optimizer over $\alpha\in[0,1]$, initialized at $\alpha=0.5$.

For inference, we follow the bootstrap procedure in \citet{liang2025quantifying}. For each inference group, we first compute $\log L_H(S_i)$ and $\log L_A(S_i)$ for all observed sentences. We then draw $B=1000$ bootstrap samples of sentences, with replacement. For each bootstrap sample $\mathcal{B}_b$, we re-estimate
\[
\hat{\alpha}^{MLE,b}
=
\arg\max_{\alpha\in[0,1]}
\frac{1}{N}\sum_{i\in\mathcal{B}_b}
\log\left[
(1-\alpha)
+
\alpha
\exp\left\{
\log L_A(S_i)-\log L_H(S_i)
\right\}
\right],
\]
retaining estimates for which the numerical optimization converges. We compute the 2.5th and 97.5th percentiles of the bootstrap estimates and report the midpoint of this interval as the point estimate, 
\[
\hat{\alpha}^{MLE}_{reported}
=
\frac{
q_{0.025}(\hat{\alpha}^{MLE,b})
+
q_{0.975}(\hat{\alpha}^{MLE,b})
}{2},
\]
with half of the interval reported as the confidence interval half-width, i.e., 
\[
CI_{half}
=
\frac{
q_{0.975}(\hat{\alpha}^{MLE,b})
-
q_{0.025}(\hat{\alpha}^{MLE,b})
}{2}.
\]

We apply the estimator separately to each country-field-quarter cell under the two benchmark choices defined above. Each quarter combines the standard calendar-quarter months.
The difference between the pooled-benchmark and country-field-specific estimates captures how much the estimated mixture weight changes when the benchmark is allowed to reflect country-field-specific writing baselines rather than imposing a common pooled baseline across all groups.

\subsection{Simulation Design}
\subsubsection{Simulated Country-field-specific log-odds Ratio}
\label{app:sim_words_ratio}
As we observe substantial dispersion in log-odds ratios across country-field groups for the top 20 AI-like words (see Appendix Table \ref{tab:variance_comparison} Column Word). We use a permutation simulation to assess whether this dispersion reflects country-field-specific writing patterns or could arise under random assignment of texts to country-field labels.

The null hypothesis is that word-use differences are unrelated to country-field membership. To construct the null distribution, we keep each publication's human abstract, AI-rewritten abstract, and publication fraction weight fixed, but randomly permute the country-field labels across rows. This preserves the overall set and size distribution of country-field groups while breaking the link between text and country-field assignment. For each permuted dataset, we recompute the country-field-specific weighted log-odds ratios using the same procedure as in the observed data. We repeat this procedure 50 times and report the results in Appendix Table \ref{tab:variance_comparison}.

\subsubsection{Simulated Pre-LLM AI-likeness}
\label{app:sim_alpha}
To assess whether the observed cross-group dispersion (Figure \ref{fig:alpha_estimation}, Full-sample Alpha bars) reflects genuine country-field structure in writing style, we conduct a permutation simulation that breaks the link between abstracts and their observed country-field labels while preserving group sizes. For each quarter of 2021, we pool all abstracts across country-field groups, randomly permute them, and partition them into synthetic country-field groups with the same sizes as in the observed data. Each simulated group, therefore, contains the same number of abstracts as its corresponding observed group, but the abstracts are randomly drawn from the pooled quarterly corpus. We then re-estimate $\alpha$ for each synthetic group using the same MLE procedure and compute the dispersion of simulated estimates across groups. We repeat this procedure 300 times, see results in Appendix Figure \ref{fig:sim_2}.

\subsection{Limitations and Future Direction}
This paper has several limitations that suggest directions for future research.

First, our analysis focuses exclusively on English-language publications. While this allows for consistent comparison across countries and fields, it may limit the generalizability of our findings to non-English scientific writing. Future work could extend the framework to multilingual settings, where baseline stylistic differences may be even more pronounced and where AI tools may play a different role.

Second, our assignment of publications to countries relies on author affiliation data, which may not precisely capture the location or identity of the individual responsible for writing the text. Our fractional and single-country assignment schemes provide reasonable approximations, and our results are robust across them, but both approaches are imperfect. Future research could leverage more granular data—such as author contribution statements or writing process metadata—to better attribute writing styles to individuals.

Third, our measurement approach is based on log-odds ratios estimated using maximum likelihood methods, which require that words appear in both human-written and AI-generated texts. As a result, words that appear exclusively in one corpus—either AI-only or human-only—are excluded from the analysis. This is a limitation, as such words may in fact provide strong signals for distinguishing between human and AI-generated writing. Future work could adopt alternative approaches, such as Bayesian estimators or smoothing-based methods, that explicitly incorporate zero-frequency features and better leverage these informative signals for classification.

Fourth, our analysis operates at the word level, abstracting from higher-order linguistic structure. While word-level measures are transparent and interpretable, they may miss more nuanced patterns in syntax and phrasing. Future work could extend this framework to n-grams, syntactic features, or embedding-based representations to capture richer dimensions of writing style.

Finally, our analysis focuses on measured AI-likeness in text, implicitly assuming that these biases may be internalized by individuals and translate into biased perceptions An important direction for future research is to use experimental approaches to elicit individuals’ real perceptions of different scientific writing. For example, randomized experiments could vary author attributes (e.g., country of affiliation or field) while holding text constant, to study how perceived ``AI-likeness,” credibility, and quality judgments respond to these labels. Such designs would help quantify how perception biases interact with stylistic features and, in turn, how they affect downstream outcomes such as citations, dissemination, and the diffusion of knowledge.

\end{document}